\documentclass[10pt,twocolumn,letterpaper]{article}

\usepackage{cvpr}
\usepackage{times}
\usepackage{epsfig}
\usepackage{graphicx}
\usepackage{amsmath}
\usepackage{amssymb}
\usepackage{subfigure}
\usepackage{stmaryrd}

\usepackage{empheq}
\usepackage{bm}
\usepackage{centernot}
\usepackage{soul}
\usepackage[subtle, title=normal, bibliography=normal]{savetrees24}

\newcommand{\bb}[1]{\textbf{#1}}
\newcommand{\mbb}[1]{\mathbb{#1}}
\newcommand{\mc}[1]{\mathcal{#1}}
\newcommand{\dd}{\mathrm{d}}

\usepackage{cancel}

\newcommand{\conj}[1]{\overline{#1}}

\usepackage[pagebackref=true,breaklinks=true,letterpaper=true,colorlinks,bookmarks=false]{hyperref}

\cvprfinalcopy % *** Uncomment this line for the final submission

\def\cvprPaperID{2077} % *** Enter the CVPR Paper ID here

\ifcvprfinal\fi
\begin{document}

%%%%%%%%% TITLE
\title{Harmonic Networks: Deep Translation and Rotation Equivariance}

\author{Daniel E.~Worrall, Stephan J.~Garbin, Daniyar Turmukhambetov and Gabriel J.~Brostow \\
{\tt \small \{{d.worrall}, {s.garbin}, {d.turmukhambetov}, {g.brostow}\}{@cs.ucl.ac.uk}}\\
University College London\thanks{\texttt{http://visual.cs.ucl.ac.uk/pubs/harmonicNets/}}}

\maketitle
%\thispagestyle{empty}

%%%%%%%%% ABSTRACT
\begin{abstract}
	Translating or rotating an input image should not affect the results 
    of many computer vision tasks. Convolutional neural networks (CNNs) 
    are already translation equivariant: input image translations produce 
    proportionate feature map translations. This is not the case for 
    rotations. Global rotation equivariance is typically sought through 
    data augmentation, but patch-wise equivariance is more difficult. 
    We present Harmonic Networks or H-Nets, a CNN exhibiting equivariance 
    to patch-wise translation and 360-rotation. We achieve this by replacing 
    regular CNN filters with circular harmonics, returning a maximal 
    response and orientation for every receptive field patch.
    
    H-Nets use a rich, parameter-efficient and fixed computational complexity 
    representation, and we show that deep feature maps within the network 
    encode complicated rotational invariants. We demonstrate that our layers 
    are general enough to be used in conjunction with the latest architectures 
    and techniques, such as deep supervision and batch normalization. We also 
    achieve state-of-the-art classification on rotated-MNIST, and competitive 
    results on other benchmark challenges.
\end{abstract}

%%%%%%%%% BODY TEXT

\section{Introduction}
\begin{figure}[t]
	\includegraphics[width=\linewidth]{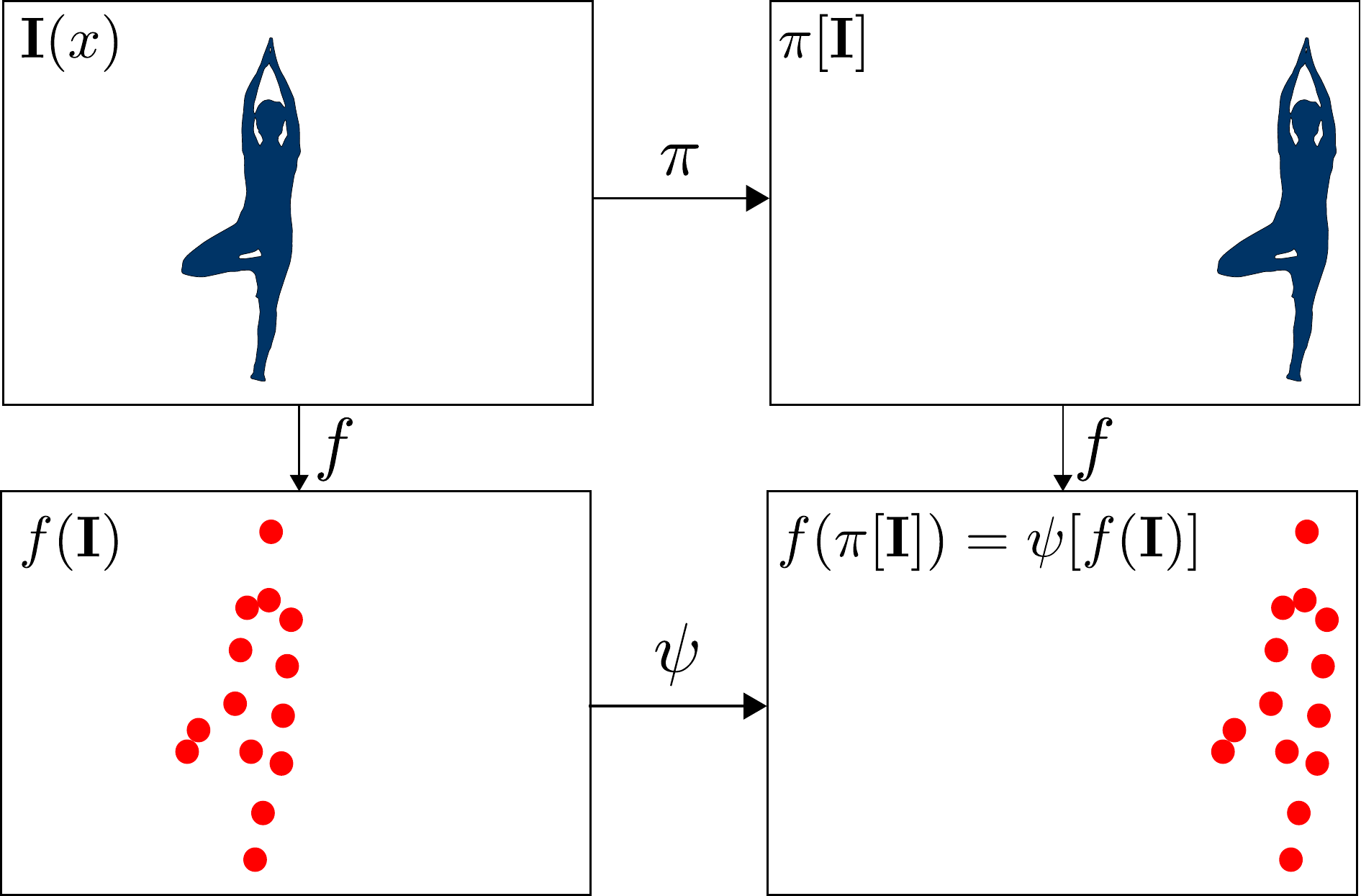}
    \caption{Patch-wise translation equivariance in CNNs arises from 
    translational weight tying, so that a translation $\pi$ of the input image 
    $\bb{I}$, leads to a corresponding translation $\psi$ of the feature 
    maps $f(\bb{I})$, where $\pi\neq \psi$ in general, due to pooling 
    effects. However, for rotations, CNNs do not yet have a feature space transformation
    $\psi$ `hard-baked' into their structure, and it is complicated to discover 
    what $\psi$ may be, if it exists at all. Harmonic Networks have
    a hard-baked representation, which allows for easier interpretation of
    feature maps---see Figure \ref{fig:commutativity}.}
    \label{fig:Z_equivariance}
\vspace{-1em}
\end{figure}

We tackle the challenge of representing $360^\circ$-rotations in convolutional neural networks (CNNs) \cite{lecun1995cnn}. Currently, convolutional layers are constrained by design to map an image to a feature vector, and \emph{translated} versions of the image map to proportionally-translated versions of the same feature vector \cite{lenc2015equivariance} (ignoring edge effects)---see Figure~\ref{fig:Z_equivariance}. However, until now, if one \emph{rotates} the CNN input, then the feature vectors do not necessarily rotate in a meaningful or easy to predict manner. The sought-after property, directly relating input transformations to feature vector transformations, is called \emph{equivariance}.

A special case of equivariance is invariance, where feature vectors remain constant under all transformations of the input. This can be a desirable property globally for a model, such as a classifier, but we should be careful not to restrict all intermediate levels of processing to be transformation invariant. For example, consider detecting a deformable object, such as a butterfly. The pose of the wings is limited in range, and so there are only certain poses our detector should normally see. A transformation invariant detector, good at detecting wings, would detect them whether they were bigger, further apart, rotated, etc., and it would encode all these cases with the same representation. It would fail to notice nonsense situations, however, such as a butterfly with wings rotated past the usual range, because it has thrown that extra pose information away. An equivariant detector, on the other hand, does not dispose of local pose information, and so it hands on a richer and more useful representation to downstream processes.
Equivariance conveys more information about an input to downstream processes, it also constrains the space of possible learned models to those that are valid under the rules of natural image formation \cite{soatto2009actionable}. This makes learning more reliable and helps with generalization. For instance, consider CNNs. The key insight is that the statistics of 
natural images, embodied in the correlations between pixels, are 
a) invariant to translation, and b) highly localized. Thus 
features at every layer in a CNN are computed on local receptive 
fields, where weights are shared across translated receptive fields. 
This weight-tying serves both as a constraint on the translational 
structure of image statistics, and as an effective technique to reduce 
the number of learnable parameters---see Figure \ref{fig:Z_equivariance}.
In essence, translational equivariance has been `baked' into the
architecture of existing CNN models. We do the same for rotation and 
refer to it as \emph{hard-baking}.

The current widely accepted practice to cope with rotation is to train with aggressive data augmentation \cite{krizhevsky2012imagenet}. This certainly improves generalization, but is not exact, fails to capture 
local equivariances, and does not ensure equivariance at every
layer within a network. How to maintain the richness of local rotation information, is what we present in this paper. Another disadvantage of data augmentation is that it leads to 
the so-called \emph{black-box} problem, where there is a lack 
of feature map interpretability. Indeed, close inspection of
first-layer weights in a CNN reveals that many of them are rotated,
scaled, and translated copies of one another \cite{zeiler2014understanding}. 
Why waste computation learning all of these redundant weights?

% Following the developments in computer vision over the last decade, 
% it has been well documented \hl{cite} that learning features instead 
% of handcrafting them, has been a boon to our field. However, even with
% a near-infinite stream of data, a 100\% learning-based approach is not 
% always the most sensible course of action. We demonstrate that handcrafting
% structure into features, on top of which we can apply learning is a 
% worthwhile and prudent way to improve performance. 

% Current CNNs are a prime example of how `hard-baking' structure into a representation 
% is beneficial for learning. 

In this paper, we present \emph{Harmonic Networks}, or \emph{H-Nets}.
They design patch-wise $360^\circ$-rotational equivariance into 
deep image representations, by constraining the filters to the family 
of \emph{circular harmonics}. The circular harmonics are \emph{steerable 
filters} \cite{freeman1991design}, which means that we can represent 
all rotated versions of a filter, using just a finite, linear combination 
of \emph{steering bases}. This overcomes the issue of learning multiple 
filter copies in CNNs, guarantees rotational equivariance, and produces
feature maps that transform predictably under input rotation.

\section{Related Work}
Multiple existing approaches seek to encode rotational equivariance into
CNNs. Many of these follow a broad approach of introducing filter or
feature map copies at different rotations. None has dominated as standard practice.

\textbf{Steerable filters}
At the root of H-Nets lies the property of \emph{filter steerability} 
\cite{freeman1991design}. Filters exhibiting steerability can be 
constructed at any rotation as a finite, linear combination of base 
filters. This removes the need to learn multiple filters at different 
rotations, and has the bonus of constant memory requirements. As such,
H-Nets could be thought of as using an infinite bank of rotated filter
copies. A work, which combines steerable filters with learning is
\cite{liu2012equi}. They build shallow features from steerable
filters, which are fed into a kernel SVM for object detection and rigid 
pose regression. H-Nets use the same filters with an added rotation offset
term, so that filters in different layers can have orientation-selectivity 
relative to one another. 

\textbf{Hard-baked transformations in CNNs}
While H-Nets hard-bake patch-wise $360^\circ$-rotation into the feature 
representation, numerous related works have encoded equivariance to 
discrete rotations. The following works can be grouped into those,
which encode global equivariance versus patch-wise equivariance, and 
those which rotate filters versus feature maps.

\cite{cohen2016group} introduce equivariance to $90^\circ$-rotations and dihedral flips in CNNs by copying the transformed filters at different rotation--flip combinations. More recently they generalized this theory to all group-structured transformations in \cite{cohen2016steer}, but they only demonstrated applications on finite groups---an extension to continuous transformations would require a treatment on anti-aliasing and bandlimiting. \cite{marcos2016learning} use a larger number of rotations for texture classification and \cite{oyallon2015scattering} also use many rotated handcrafted filter copies, opting not to learn the filters. To achieve equivariance to a greater number of rotations, these methods would need an infinite amount of computation. H-Nets achieve equivariance to all rotations, but with finite computation. 

\cite{fasel2006neocognitron} feed in multiple rotated copies of the CNN input and fuse the output predictions. \cite{laptev2016tipooling} do the same for a broader class of global image transformations, and propose a novel per-pixel pooling technique for output fusion. As discussed, these techniques lead to global equivariances only and do not produce interpretable feature maps. \cite{dieleman2016exploiting} go one step further and copy each feature map at four $90^\circ$-rotations. They propose 4 different equivariance preserving feature map transformations. Their CNN is similar to \cite{cohen2016group} in terms of what is being computed, but rotating feature maps instead of filters. A downside of this is that all inputs and feature maps have to be square; whereas, we can use any sized input.

\textbf{Learning generalized transformations}
Others have tried to learn the transformations directly from the data. While this is an appealing idea, as we have said, for certain transformations it makes more sense to hard-bake these in for interpretability and reliability. \cite{memisevic2010hobm} construct a higher-order Boltzmann machine, which learns tuples of transformed linear filters in input--output pairs. Although powerful, they have only shown this to work on shallow architectures. \cite{hinton2011transforming} introduced \emph{capsules}, units of neurons designed to mimic the action of cortical columns. Capsules are designed to be invariant to complicated transformations of the input. Their outputs are merged at the deepest layer, and so are only invariant to global transformation. \cite{lenc2016covariant} present a method to regress equivariant feature detectors using an objective, which penalizes representations, which lie far from the equivariant manifold. Again, this only encourages global equivariance; although, this work could be adapted to encourage equivariance at every layer of a deep pipeline.

\section{Problem analysis}
Many computer vision systems strive to be view independent, such as object 
recognition, which is invariant to affine transformations, or boundary detection, 
which is equivariant to non-rigid deformations. H-Nets hard-bake $360^\circ$-rotation
equivariance into their feature representation, by constraining the convolutional 
filters of a CNN to be from the family of circular harmonics. Below, we outline 
the formal definition of equivariance (Section \ref{sec:equivariance}), how the 
circular harmonics exhibit rotational equivariance (Section \ref{sec:circular_harmonics}) 
and some properties of the circular harmonics, which we must heed for successful
integration into the CNN framework (Section \ref{sec:properties}).

\textbf{Continuous domain feature maps}
In deep learning we use feature maps, which live in a discrete domain.
We shall instead use continuous spaces, because the analysis
is easier. Later on in Section \ref{sec:sampling} we shall demonstrate
how to convert back to the discrete domain for practical implementation,
but for now we work entirely in continuous Euclidean space.

\subsection{Equivariance}
\label{sec:equivariance}
Equivariance is a useful property to have because transformations 
$\pi$ of the input produce predictable transformations $\psi$ of 
the features, which are interpretable and can make learning easier.
Formally, we say that feature mapping $f : \mc{X} \to \mc{Y}$ is \emph{equivariant} 
to a group of transformations if we can associate every transformation
$\pi\in\Pi$ of the input $\bb{x}\in\mc{X}$ with a transformation 
$\psi\in \Psi$ of the features; that is,
\begin{align}
	\psi[f(\bb{x})] = f(\pi[\bb{x}]) \label{eq:equivariance}.
\end{align} 
This means that the order, in which we apply the feature mapping and the 
transformation is unimportant---they \emph{commute}. An example is depicted 
in Figure \ref{fig:Z_equivariance}, which shows that in CNNs the order of 
application of integer pixel-translations and the feature map are interchangeable.
An important point of note is that $\pi \neq \psi$ in general, so if we 
seek for $\Pi$ to be rotations in the image domain, we do not require 
to find the set of $f$, such that $\Psi$ ``looks like'' a rotation in 
feature space, rather we are searching for the set of $f$, such that there 
exists an \emph{equivalent} class of transformations $\Psi$ in feature 
space. A special case of equivariance is \emph{invariance}, when 
$\Psi=\{\mbb{I}\}$, the identity.
\begin{figure}[t]
	\includegraphics[width=\linewidth]{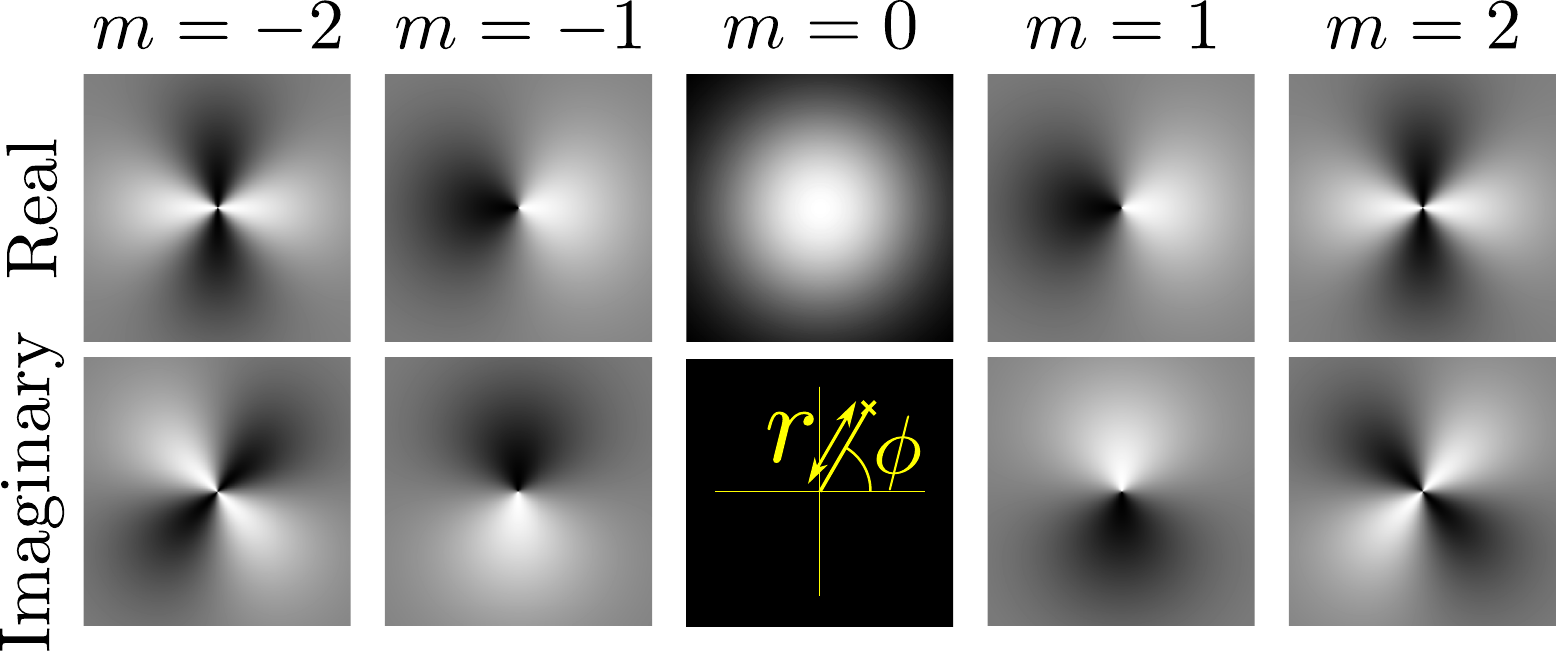}
    \caption{Real and imaginary parts of the complex Gaussian filter 
    $\bb{W}_m(r,\phi';e^{-r^2},0)=e^{-r^2}e^{im\phi}$, for
    some rotation orders. As a simple example, we have set $R(r)=e^{-r^2}$
    and $\beta=0$, but in general we learn these quantities. 
    Cross-correlation, of a feature map 
    of rotation order $n$ with one of these filters of rotation order 
    $m$, results in a feature map of rotation order $m+n$. Note the negative rotation order filters have flipped imaginary parts compared to the positive orders.}
    \label{fig:circular_harmonics}
\vspace{-1em}
\end{figure}
\subsection{The Complex Circular Harmonics}
\label{sec:circular_harmonics}
With data augmentation CNNs may learn some rotation equivariance, but this is difficult to quantify \cite{lenc2015equivariance}. H-Nets take the simpler approach of hard-baking this structure in. If $f$ is the feature mapping of a standard convolutional layer, then $360^\circ$-rotational equivariance can be hard-baked in by restricting the filters to be of the from the circular harmonic family (proof in Supplementary Material)
\begin{align}
	\bb{W}_m(r,\phi;R,\beta) = R(r)e^{i(m\phi + \beta)} \label{eq:hnet_filters}.
\end{align}
Here $r,\phi$ are the spatial coordinates of image/feature maps, expressed in polar form, $m\in\mbb{Z}$ is known as the \emph{rotation order}, $R:\mbb{R}_+\to\mbb{R}$ is a function, called the \emph{radial profile}, which controls the overall shape of the filter, and $\beta\in[0,2\pi)$ is a \emph{phase offset} term, which gives the filter orientation-selectivity. During training, we learn the radial profile and phase offset terms. Examples of the real component of $\bb{W}_m$ for a `Gaussian envelope' and different rotation orders are shown in Figure \ref{fig:circular_harmonics}. Since we are dealing with complex-valued filters, all filter responses are complex-valued, and we assume from now on that the reader understands that all feature maps are complex-valued, unless otherwise specified. Note that there are other works (e.g.,\ \cite{tygert2016complex}), which use complex filters, but our treatment differs in that the complex phase of the response is explicitly tied to rotation angle.
\begin{figure}[t]
	\includegraphics[width=\linewidth]{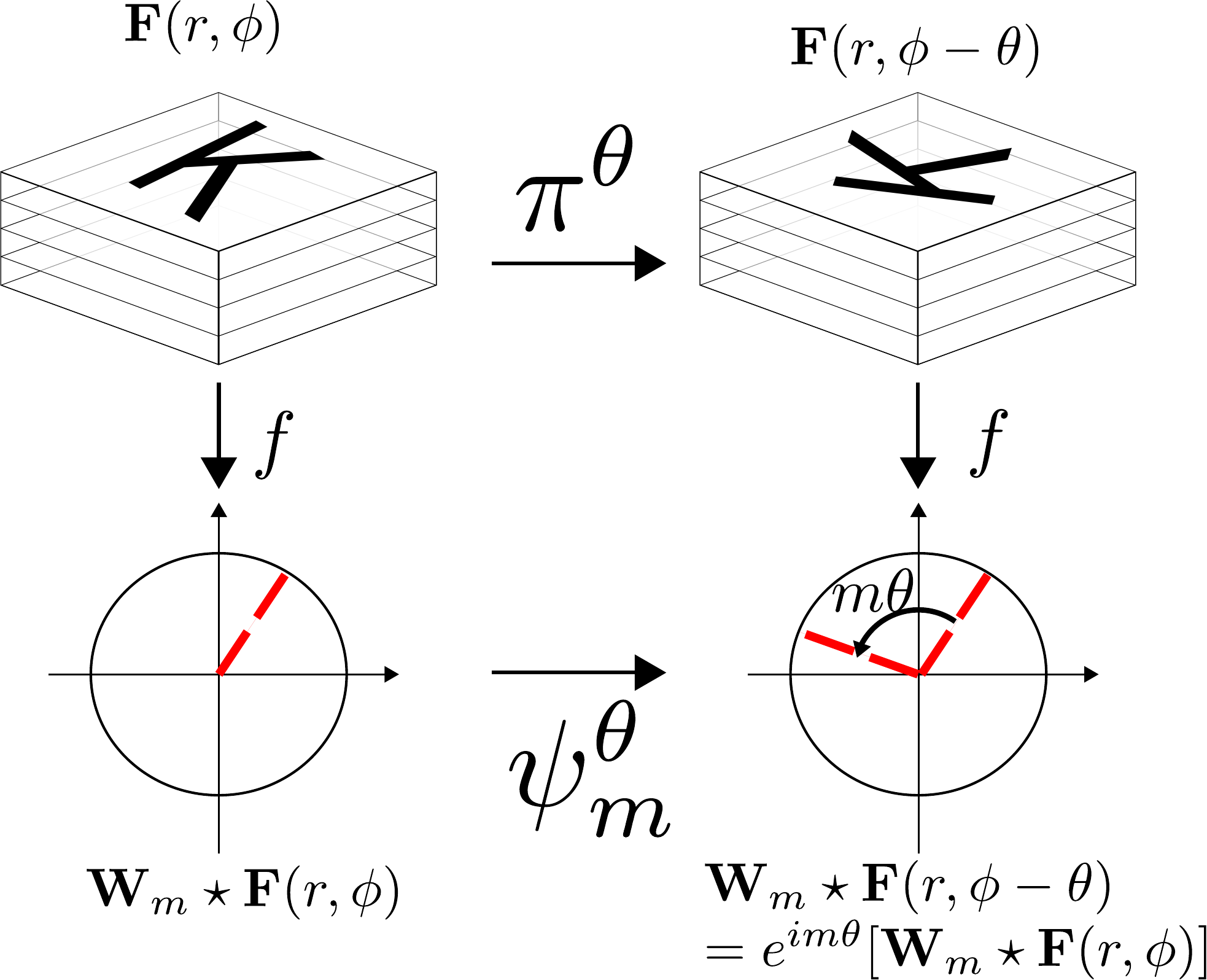}
    \caption{\textsc{Down}: Cross-correlation of the input patch with $\bb{W}_m$ yields a scalar complex-valued response. 
    \textsc{Across-then-down}: Cross-correlation with the $\theta$-rotated 
    image yields another complex-valued response. \textsc{Bottom}: 
    We transform from the unrotated response to the rotated response, through
    multiplication by $e^{im\theta}$.}
    \label{fig:commutativity}
\vspace{-1em}
\end{figure}

\textbf{Rotational Equivariance of the Circular Harmonics}
Some deep learning libraries implement cross-correlation $\star$ rather than convolution $*$, and since the understanding is slightly easier to follow, we consider correlation. Strictly, cross-correlation with complex functions requires that one of the arguments is conjugated, but we do not do this in our model/implementation, so
\begin{align}
	[\textbf{W}\star\textbf{F}](p',q') &= \int \textbf{W}(p-p',q-q')\textbf{F}(p,q) \, \dd p \dd q \\
    [\textbf{W}*\textbf{F}](p',q') &= \int \textbf{W}(p'-p,q'-q)\textbf{F}(p,q)  \, \dd p \dd q.
\end{align}
Consider correlating a circular harmonic of order $m$ with a rotated image patch. We assume that the image patch is only able to rotate locally about the origin of the filter. This means that the cross-correlation response is a scalar function of input image patch rotation $\theta$. Using the notation from Equation \ref{eq:equivariance}, and recalling that we are working in polar coordinates $(r,\phi)$, counter-clockwise rotation of an image $\bb{F}(r,\phi)$ about the origin by an angle $\theta$ is $\bb{F}(r,\pi^\theta [\phi]) = \bb{F}(r,\phi - \theta)$. As a shorthand we denote $\bb{F}^{\theta} := \bb{F}(r,\pi^{\theta}[\phi])$. It is a well-known result \cite{liu2012equi,freeman1991design} (proof in Supplementary Material) that
\begin{align}
	[\bb{W}_m &\star \bb{F}^{\theta}] = e^{im\theta} [\bb{W}_m \star \bb{F}^0], \label{eq:new_convolution}
\end{align}
where we have written $\bb{W}_m$ in place of $\bb{W}_m(r,\phi;R,\beta)$ for brevity. We see that the response to a $\theta$-rotated image $\bb{F}^{\theta}$ with a circular harmonic of order $m$ is equivalent to the cross-correlation of the unrotated image $\bb{F}^0$ with the harmonic, followed by multiplication by $e^{im\theta}$. While the rotation is done in input space,  multiplication by $e^{im\theta}$ is performed in feature space, and so, using the notation from Equation 1, $\psi_m^\theta[\bullet] = e^{im\theta}\cdot \bullet$. This process is shown in Figure \ref{fig:commutativity}. Note that we have included a subscript $m$ on the feature space transformation. This is important, because the kind of feature space transformation we apply is dependent on the rotation order of the harmonic. Because the phase of the response rotates with the input at frequency $m$, we say that the response is an \emph{$m$-equivariant feature map}. By thinking of an input image as a complex-valued feature map with zero imaginary part, we could think of it as $0$-equivariant. 

The rotation order of a filter defines its response properties to input rotation. In particular, rotation order $m=0$ defines invariance and $m=1$ defines linear equivariance. For $m=0$ this is because, denoting $f_m := [\bb{W}_m\star\bb{F}^0]$, then $\psi_0^\theta [f_m] = e^{i\cdot 0\theta}\cdot f_m = f_m$, which is independent of $\theta$. For $m=1$, $\psi_1^\theta [f_m] = e^{i\cdot 1\theta} f_m$---as the input rotates, $e^{i\theta} f_m$ is a complex-valued number of constant magnitude $f_m$, spinning round with a phase equal to $\theta$. Naturally, we are not constrained to using rotation orders 0 or 1 only, and we make use of higher and negative orders in our work. 

\textbf{Arithmetic and the Equivariance Condition}
\label{sec:properties}
Further important properties of the circular harmonics, which are proven 
in the Supplementary Material, are: 1) Chained cross-correlation of rotation 
orders $m_1$ and $m_2$ lead to a new response with rotation order $m_1+m_2$.
2) Point-wise nonlinearities $h : \mbb{C} \to \mbb{C}$, acting solely on the 
magnitudes maintain rotational equivariance, so we can interleave 
cross-correlations with typical CNN nonlinearities adapted to the complex 
domain. 3) The summation of two responses of the same order $m$ remains of 
order $m$. Thus to construct a CNN where the output is $M$-equivariant 
to the input rotation, we require that the sum of rotation orders along 
any path equals $M$, so
\begin{align}
	\sum_{i=1}^N m_{i} = M.
\end{align}
This is the fundamental condition underpinning the equivariance properties
of H-Net, so we call it the \emph{equivariance condition}.

We note here that for our purposes, our filter $\bb{W}_{-m} = \conj{\bb{W}_m}$ (the 
complex conjugate), which saves on parameters, but this does not necessarily 
imply conjugacy of the responses unless $\bb{F}$ is real, which is only true 
at the input.

\section{Method}
We have considered the $360^\circ$-rotational equivariance of feature maps
arising from cross-correlation with the circular harmonics, and we determined 
that the rotation orders of chained cross-correlations sum. Next, we use these 
results to construct a deep architecture, which can leverage the equivariance 
properties of circular harmonics.

\subsection{Harmonic Networks}
\begin{figure}[t]
	\includegraphics[width=\linewidth]{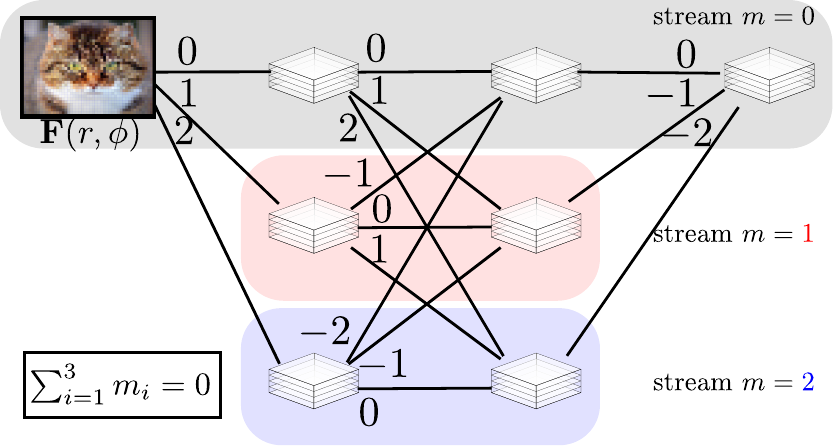}
    \caption{An example of a 2 hidden layer H-Net with $m=0$ output, 
    input--output left-to-right. Each horizontal stream represents a series 
    of feature maps (circles) of constant rotation order. The edges 
    represent cross-correlations and are numbered with the rotation 
    order of the corresponding filter. The sum of rotation orders along
    any path of consecutive edges through the network must equal $M=0$,
    to maintain disentanglement of rotation orders.}
    \label{fig:H-net_example}
\vspace{-1em}
\end{figure}
The rotation order of feature maps and filters sum upon cross-correlation, so to achieve a given output rotation order, we must obey the equivariance condition. In fact, at every feature map, the equivariance condition must be met, otherwise, it should be possible to arrive at the same feature map along two different paths, with different summed rotation orders. The problem is that combining complex features, with phases, which rotate at different frequencies, leads to \emph{entanglement} of the responses. The resultant feature map is no longer equivariant to a single rotation order, making it difficult to work with. We resolve this by enforcing the equivariance condition at every feature map.

Our solution is to create separate streams of constant rotation order responses running through the network---see Figure \ref{fig:H-net_example}. These streams contain multiple layers of feature maps, separated by rotation order zero cross-correlations and nonlinearities. Moving between streams, we use cross-correlations of rotation order equal to the difference between those two streams. It is very easy to check that the equivariance condition holds in these networks.

When multiple responses converge at a feature map, we have multiple choices of how to combine them. We could stack them, we could pool across them, or we could sum them \cite{dieleman2016exploiting}. To save on memory, we chose to sum responses of the same rotation order
\begin{align}
	\bb{Y}_p = \sum_{m,n:m+n=p} \bb{W}_m \star \bb{F}_n.
\end{align}
$\bb{Y}_p$ is then fed into the next layer. Usually in our experiments, 
we use streams of orders 0 and 1, which we found to work well and is justified by the fact that CNN filters tend to contain very little high frequency information \cite{jacobsen2016structure}.
\begin{figure}[t]
	\centering
	\includegraphics[width=0.7\linewidth]{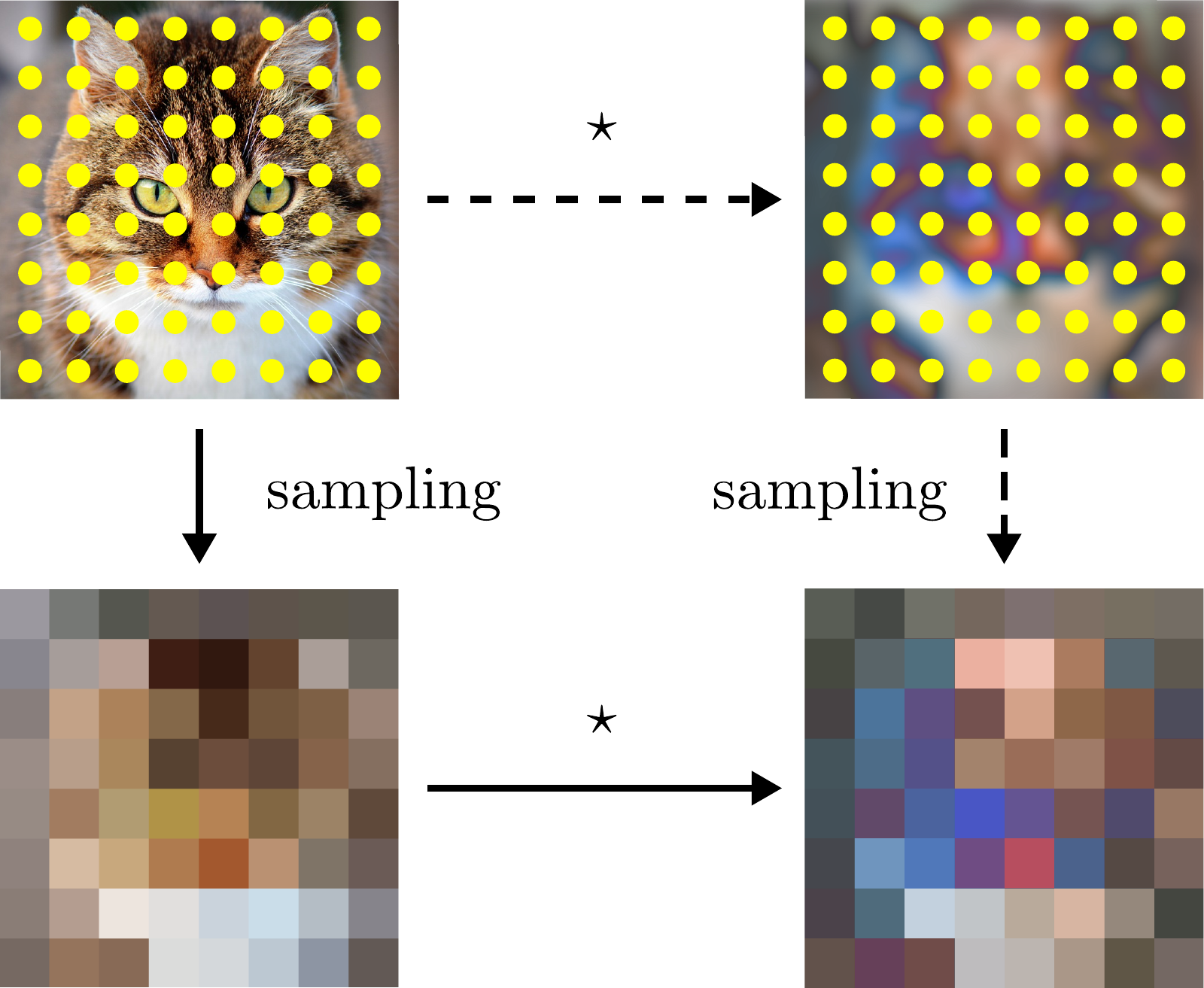}
    \caption{H-Nets operate in a continuous spatial domain, but we can
    implement them on pixel-domain data because sampling and 
    cross-correlation commute. The schematic shows an example of 
    a layer of an H-Net (magnitudes only). The solid arrows follow 
    the path of the implementation, while the dashed arrows follow 
    the possible alternative, which is easier to analyze, but 
    computationally infeasible. The introduction
    of the sampling defines \emph{centers of equivariance} at 
    pixel centers (yellow dots), about which a feature map is 
    rotationally equivariant. }
    \label{fig:sampling_grid}
\vspace{-1em}
\end{figure}

Above, we see that the structure of the Harmonic Network is very
simple. We replaced regular CNN filters with radially reweighted and
phase shifted circular harmonics. This causes each filter response to be
equivariant to input rotations with order $m$. To prevent responses of 
different rotation order from entangling upon summation, we separated
filter responses into streams of equal rotation order.

\textbf{Complex nonlinearities}
\label{sec:complex_nonlinearities}
Between cross-correlations, we use complex nonlinearities, which act on the 
magnitudes of the complex feature maps only, to preserve rotational equivariance.
An example is a complex version of the ReLU
\begin{align}
	\mbb{C}\text{-ReLU}_b(Xe^{i\phi}) = \text{ReLU}(X+b)e^{i\phi}.
\end{align}
We can provide similar analogues for other nonlinearities and for Batch 
Normalization \cite{ioffe2015bn}, which we use in our experiments.

We have thus far presented the Harmonic Network. Each layer is a collection
of feature maps of different rotation orders, which transform predictably
under rotation of the input to the network and the $360^\circ$-rotation 
equivariance is achieved with finite computation. Next we show how to 
implement this in practice.

\subsection{Implementation: Discrete cross-correlations}
\label{sec:sampling}
Until now, we have operated on a domain with continuous spatial dimensions $\Omega=\mbb{R}\times\mbb{R}\times\{1,k_{\ell}\}$. However, the H-Net needs to operate on real-world images, which are sampled on a 2D-grid, thus we need to anti-alias the input to each discretized layer. We do this with a simple Gaussian blur. We can then use a regular CNN architecture without any problems. This works on the fact that the order of bandlimited sampling and cross-correlation is interchangeable~\cite{freeman1991design}; so either we correlate in continuous space, then downsample, or downsample then correlate in the discrete space. Since point-wise nonlinearities and sampling also commute, the entire H-Net, seen as a deep feature-mapping, commutes with sampling. This could allow us to implement the H-Net on non-regular grids; although, we did not explore this. 

Viewing cross-correlation on discrete domains sheds some insight into how the equivariance properties behave. In Figure \ref{fig:sampling_grid}, we see that the sampling strategy introduces multiple origins, one for each feature map patch. We call these, \emph{centers of equivariance}, because a feature map will exhibit local rotation equivariance about each of these points. If we move to using more exotic sampling strategies, such as strided cross-correlation or average pooling, then the centers of equivariance are ablated or shifted. If we were to use max-pooling, then the center of equivariance would be a complicated nonlinear function of the input image and harmonic weights. For this reason we have not used max-pooling in our experiments.

\textbf{Complex cross-correlations}
\begin{figure}[t]
\begin{center}
	\includegraphics[width=\linewidth]{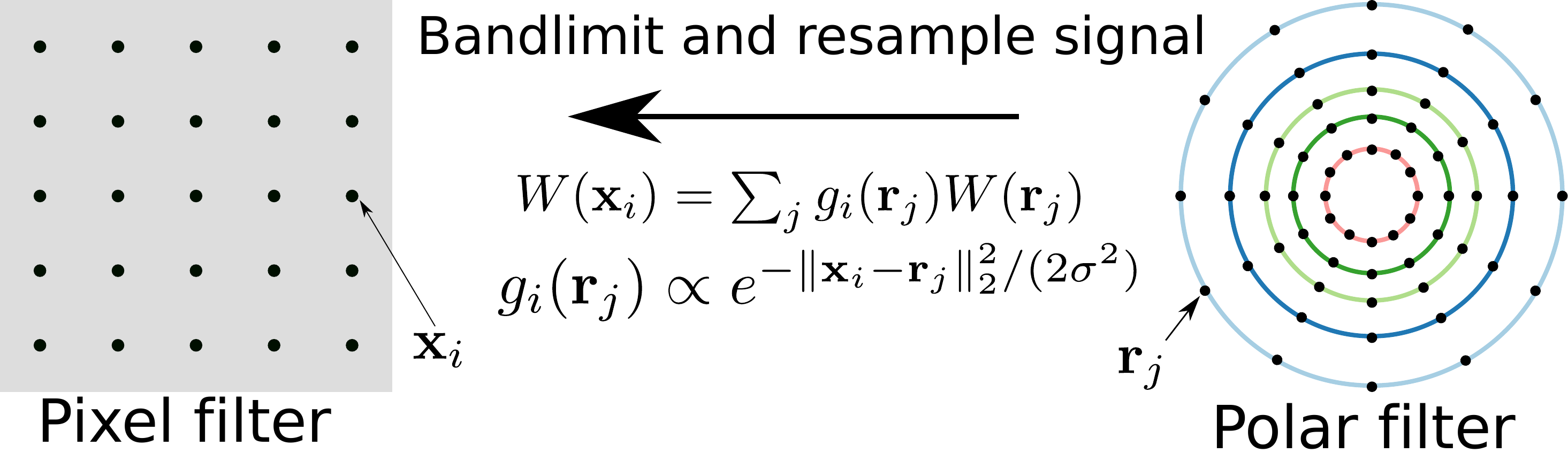}
\end{center}
\vspace{-1em}
\caption{Images are sampled on a rectangular grid but our filters are defined in the polar domain, so we bandlimit and resample the data before cross-correlation via Gaussian resampling.}
\label{fig:weights}
\vspace{-1em}
\end{figure}
On a practical note, it is worth mentioning, that complex cross-correlation can be implemented efficiently using 4 real cross-correlations
\begin{align}
	\underbrace{\bb{W}_m^{\text{Re}}\star\bb{F}^{\text{Re}} - \bb{W}_m^{\text{Im}}\star\bb{F}^{\text{Im}}}_{\text{real response}} + i \underbrace{\bb{W}_m^{\text{Re}}\star\bb{F}^{\text{Im}} + \bb{W}_m^{\text{Im}}\star\bb{F}^{\text{Re}})}_{\text{imaginary response}}.
\end{align}
So circular harmonics can be implemented in current deep learning frameworks, with minor engineering. We implement a grid-resampled version of the filters $W(\bb{x}_i)=\sum_j g_i(\bb{r}_j)W(\bb{r}_j)$, with $g_i(\bb{x}_j)\propto e^{-\|\bb{r}_i-\bb{x}_j\|_2^2/(2\sigma^2)}$ (see Figure \ref{fig:weights}). The polar representation $(r_j,\phi_j)$ can be mapped from the components $\bb{r}_j$ by $\bb{r}_j = [r_j\cos \phi_j ,\, r_j\sin \phi_j]^\top$. If we stack all the polar filter samples into a matrix we can write each point as the outer product of a radial tensor $\bb{R}_j$ and trigonometric angular tensor $[\cos m\bm{\Phi}_{r_j}, i \sin m\bm{\Phi}_{r_j}]^\top$. The phase offset $\beta$ can be separated out by noting that
\begin{align}
W_m(\bb{r}_j) = \sum_{i=1}^I R(r_j) \begin{bmatrix}
	\bb{I}\cos \beta	&	-\bb{I}\sin \beta \\
	\bb{I}\sin \beta	&	\bb{I}\cos \beta
	\end{bmatrix} \begin{bmatrix}
	\cos m\bm{\Phi}_{r_j} \\
	i \sin m\bm{\Phi}_{r_j}
	\end{bmatrix}
\end{align}
where the complex exponential and trigonometric terms are element-wise, 
and $\bb{I}$ is the identity matrix. This is just a reweighting of the 
ring elements. In full generality, we could also use a per-radius phase 
$\beta_{r_i}$, which would allow for spiral-like left- and right-handed 
features, but we did not investigate this.

\subsection{Computational cost}
We have increased the computational cost of cross-correlation, in return for continuous rotational equivariance. Here we analyze the computational cost in terms of number of multiplications.
\begin{figure}[t]
	\includegraphics[width=\linewidth]{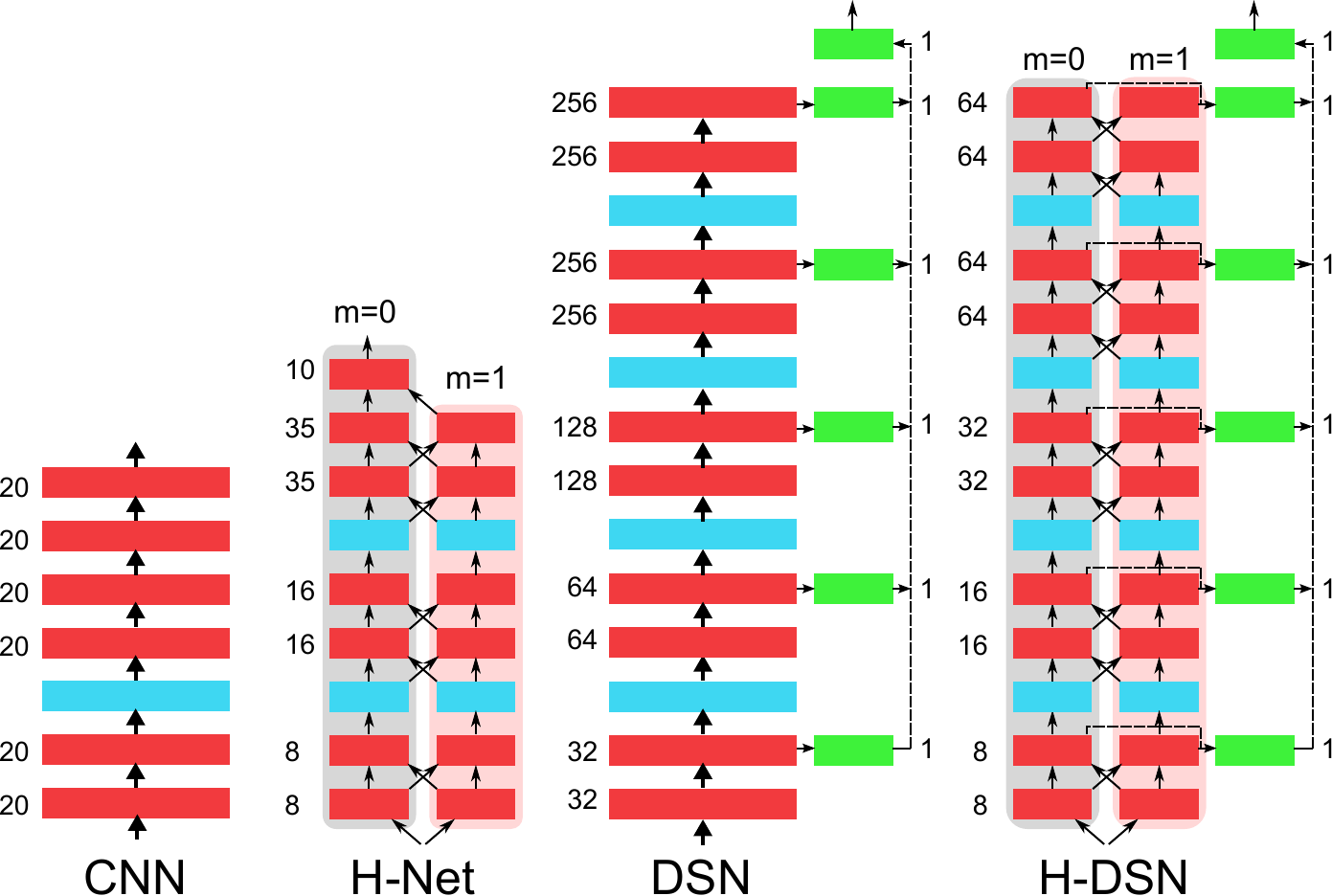}
    \caption{Networks used in our experiments. \textsc{Left}:
    MNIST networks, as per \cite{cohen2016group}. \textsc{Right} 
    deeply-supervised networks (DSN) \cite{lee2015deeply} 
    for boundary segmentation, as per \cite{xie2015hed}. Red
    boxes denote feature maps. Blue boxes are pooling (max for CNNs and
    average for H-Nets). Green boxes are side feature maps
    as per \cite{xie2015hed}; these are connected to the DSN with dashed
    lines for ease of viewing. All main cross-correlations are $3\times3$,
    unless otherwise stated in the experiments section.}
    \label{fig:networks}
\vspace{-1em}
\end{figure}
In the standard cross-correlation, for an input of size $h\cdot w\cdot i_{\mbb{Z}}$, (height, width, input channels) and filters of size $k\cdot k\cdot o_{\mbb{Z}}$ (height, width, output channels), the number of multiplications to form a feature map of the same size as the input is $M(\mbb{Z})=hwk^2i_{\mbb{Z}}o_{\mbb{Z}}$. In the H-Net, we have $f$ rotation orders on the input and $r$ rotation orders on the output, so perform $fr$ complex cross-correlations. Each complex cross-correlation can be formed from 4 real cross-correlations, so the number of multiplications is $4M(\mbb{H})fr$, where $i_{\mbb{H}}$ and $o_{\mbb{H}}$ are the number of input and output channels, respectively. Thus for similar computational cost we equate the two to yield $M(\mbb{Z}) = 4M(\mbb{H})fr$. Rearranging; setting $i_\mbb{H}=o_\mbb{H}$, $i_\mbb{Z}=o_\mbb{Z}$ and $f=r$; and taking the square root of both sides, we arrive at a simple rule of thumb for network design,
$i_{\mbb{Z}} = 2fi_{\mbb{H}}$.
For example, if we want to build an H-Net with similar computational cost to a regular CNN with 64 channels per layer, then if we use 2 rotation orders $m\in\{0,1\}$, then the number of H-Net channels is $64/(2\cdot2) = 16$.

\section{Experiments}
We validate our rotation equivariant formulation below, performing
some introspective investigations, and measuring against relevant
baselines for classification on the rotated-MNIST dataset~\cite{larochelle2007deep} 
and boundary detection on the Berkeley Segmentation Dataset~\cite{arbelaez2011bsd}.
We selected our baselines as representative examples of the current 
state-of-the-art and to demonstrate that H-Nets can be used on different 
architectures for different tasks. The networks we used are in Figure \ref{fig:networks}.

\subsection{Benchmarks}
Here we compare H-Nets for classification and boundary detection. 
Classification is a typical rotation invariant task, and should 
suit H-Nets very well. In contrast, boundary detection is a rotation 
equivariant task. The key to the success of the H-Net is that it can achieve
global equivariance, without sacrificing local equivariance of features.

\textbf{MNIST}
Of course, this is a small dataset, with simple visual structures, but it is a 
good indication of how introducing the right equivariances into CNNs can aid 
inference.
\begin{table}[t]
  \begin{center}
    {
    \begin{tabular}{|l|c|c|}
    \hline
    Method & Test error (\%) & \# params\\
    \hline\hline
    SVM \cite{larochelle2007deep}						& $11.11$		& -\\
    Transformation RBM \cite{sohn2012local}				& $4.2$ 		& -\\
    Conv-RBM \cite{schmidt2012priors} 					& $3.98$ 		& -\\
    CNN \cite{cohen2016group}							& $5.03$ 		& 22k\\
    CNN \cite{cohen2016group} + data aug*				& $3.50$ 		& 22k\\
    $P4$CNN rotation pooling \cite{cohen2016group}		& $3.21$		& 25k\\
    $P4$CNN \cite{cohen2016group} 						& $2.28$		& 25k\\
    H-Net (Ours)										& $\bb{1.69}$ 	& 33k\\
    \hline
    \end{tabular}
    }
  \end{center}
  \caption{Results. Our model sets a new state-of-the-art on the 
  rotated MNIST dataset, reducing the test error by 26\%. * Our
  reimplementation}
  \label{tab:MNIST}
\end{table}
We investigate classification on the rotated MNIST dataset (new version) \cite{larochelle2007deep} 
as a baseline. This has 10000 training images, 2000 validation images, and
50000 test images. The $360^\circ$-rotations and small training set
size make this a difficult task for classical CNNs. We compare against 
a collection of previous state-of-the-art papers and 
\cite{cohen2016group}, who build a deep CNN with filter copies at 
$90^\circ$-rotations. We try to mimic their network architecture for
H-Nets as best as we can, using 2 rotation order streams with $m\in\{0,1\}$ 
through to the deepest layer, and complex-valued versions of ReLU 
nonlinearities and Batch Normalization (see Method). We also replace
max-pooling with mean-pooling layers, as shown in Figure \ref{fig:networks}. 
We perform stochastic gradient descent on a cross-entropy loss using 
Adam \cite{kingma2014adam} and an adaptive learning rate, which we divide by 10 if there has 
been no improvement in validation accuracy in the last 10 epochs. We 
train multiple models with randomly chosen hyperparameters, and report 
the test error of the model that performed best on the validation set, 
training on a combined training and validation set Table~\ref{tab:MNIST} 
lists our results. This model actually has 33k parameters, which is about 50\%
larger than the standard CNN and \cite{cohen2016group}, which have 22k. 
This is because it uses $5\times 5$ convolutions instead of $3\times 3$.
Interestingly, it does not overfit on such a small dataset and it still
outperforms the standard CNN trained with rotation augmentations, which
we do not use. We set the new state-of-the-art, with a 26\% 
improvement on the previous best model.

\begin{table}[t]
  \begin{center}
  {
    \begin{tabular}{|l|c|c|c|}
    \hline
    Method & ODS & OIS & \# params\\
    \hline\hline
    HED, \cite{xie2015hed}*							& 0.640 		& 0.650			& 2346k \\
    HED, low \# params \cite{xie2015hed}*				& 0.697 		& 0.709			& 115k \\
    Kivinen et al. \cite{kivinen2014boundary}		& 0.702			& 0.715			& -\\
    H-Net (Ours)									& $\bb{0.726}$	& $\bb{0.742}$	& 116k\\
    \hline
    CSCNN\dag, \cite{hwang2015bsd}					& 0.741			& 0.759 & \\
    DeepEdge\dag, \cite{bertasius2015bsd}			& 0.753			& 0.772 & \\
    $N^4$-Fields\dag, \cite{ganin2014bsd}			& 0.753			& 0.769 & \\
    DeepContour\dag, \cite{shen2015bsd}				& 0.756			& 0.773	& \\
    HED\dag,  \cite{xie2015hed}				& 0.782 		& 0.804			& 2346k \\
    DCNN + sPb\dag, \cite{kokkinos2015bsd}			& $\bb{0.813}$	& $\bb{0.831}$ & \\
    \hline
    \end{tabular}
    }
  \end{center}
  \caption{Our model beats the non-pretrained neural networks baselines on BSD500~\cite{arbelaez2011bsd}. *Our implementation. \dag ImageNet pretrained}
  \label{tab:bsd}
\vspace{-1em}
\end{table}

\textbf{Deep Boundary Detection} Boundary detection is equivariant to non-rigid transformations;
although edge presence is locally invariant to orientation. The
current state-of-the-art depends on fine-tuning ImageNet-pretrained
networks to regress boundary probabilities on a per-patch basis.
To demonstrate that hard-baked rotation equivariance serves as a
strong generalization tool, we compared against a previous state-of-the-art
architecture \cite{xie2015hed}, \emph{without pretraining}. We tried 
to mimic \cite{xie2015hed} as closely as possible, as shown in Figure \ref{fig:networks}. The main difference is
that we divide the number of all feature maps by 2, for faster,
more stable training. They use a VGG network~\cite{simonyan2014vgg} 
extended with deeply supervised network (DSN)~\cite{lee2015deeply} 
\emph{side-connections}. These are $1\times 1$-convolutions, which 
perform weighted averages across all relevant feature maps, resized 
to match the input. A binary cross-entropy loss is applied to each 
side connection, to stabilize learning. A final `fusion' layer is 
created by taking a weighted linear combination of the side-connections, 
this is the final output. We adapt side-connections to H-Nets, by
using the complex magnitude of feature maps before taking a weighted
average. This means that the resultant boundary predictions are
locally invariant to rotation. We added a small sparsity regularizer
to our cost function, because we found it improved results slightly.
We call the Harmonic variant of the DSN, an \emph{H-DSN}. We also compare against \cite{xie2015hed} with the number of parameters matched to H-DSN (the first layer has 7 features, instead of 16, and so on).

We also compared with \cite{kivinen2014boundary}, who use a 
mean-and-covariance-RBM. Their technique has five main contributions:
1) zero-mean, unit variance normalization of inputs, 2) sparsity
regularization of hidden units, 3) averaged ground truth edge 
annotations, 4) average outputs to 16 input rotations, 5)
non-maximum suppression of results by the Canny method.
We perform the first 2 methods, but leave the last 3 alone.
In particular, they do not pretrain on ImageNet, and attempt some
kind of rotation averaging for global equivariance, so are a good baseline
to measure against. We tested on the Berkeley Segmentation
Dataset (BSD500)~\cite{arbelaez2011bsd}. As shown in Table~\ref{tab:bsd} 
for non-pretrained models, H-Nets deliver superior performance over 
current state-of-the-art architectures, including \cite{kivinen2014boundary}, 
who also encode rotation equivariance. Most noticeable of all is
that we only use 5\% of the parameters of \cite{xie2015hed}, showing
how by restricting the search space of learnable models through 
hard-baking local rotation equivariance, we need not learn so many
parameters.

\subsection{Model Insight}
Here we investigate some of the properties of the H-Net implementation,
making sure that the motivations behind H-Net design are achieved by
the implementation.

\begin{figure}[t]
\begin{center}
	\includegraphics[width=0.8\linewidth]{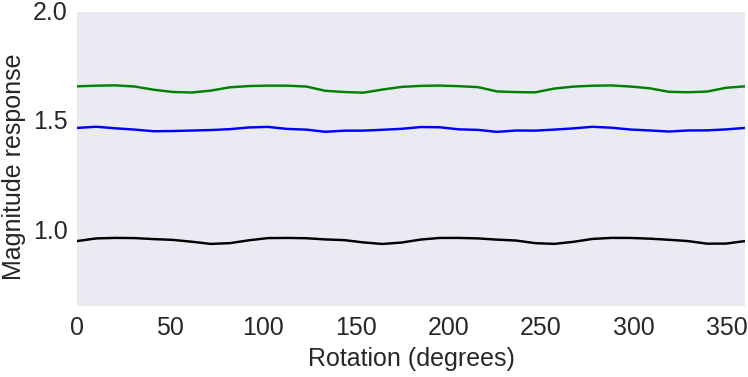}
\end{center}
\vspace{-1em}
\caption{Stability of the response magnitude to input rotation angle. Black $m=0$, blue $m=1$, green $m=2$.}
\label{fig:stability}
\vspace{-1em}
\end{figure}

\textbf{Rotational stability}
As a sanity check, we measured the invariance of the magnitude response to rotation for $m\in\{0,1,2\}$. We show the result of rotating a random input to an H-Net layer in Figure \ref{fig:stability}. The response is very flat, with periodic small fluctuations due to the inexactness of anti-aliasing.

\textbf{Filter Visualization}
\begin{figure}[b]
	\includegraphics[width=\linewidth]{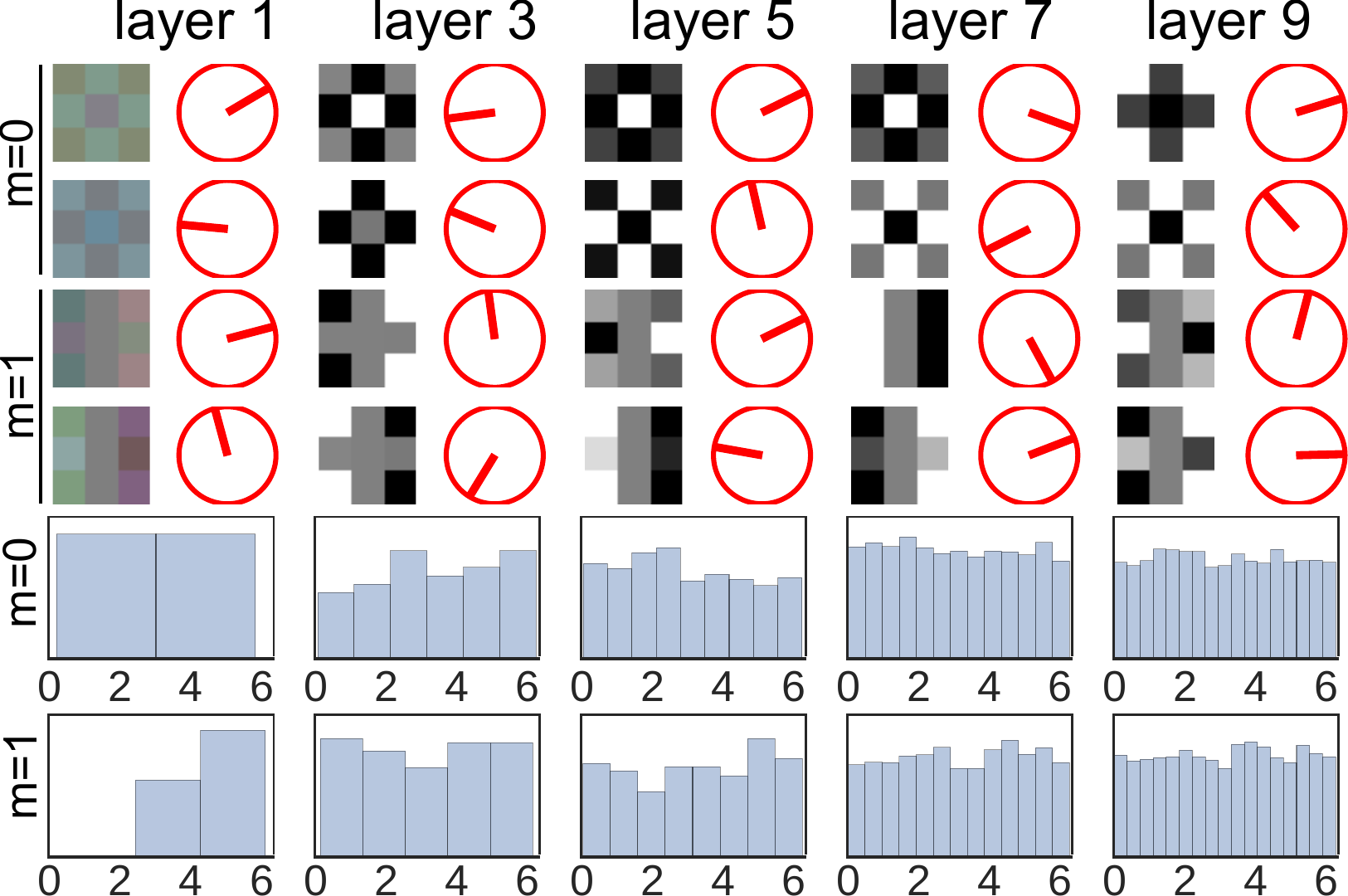}
    \caption{Randomly selected filters and phase histograms from the 
    BSDS500 trained H-DSN. Filter are aligned at $\beta=0$; and the 
    oriented circles represent phase. We see few filter copies and no
    blank filters, as usually seen in CNNs. We also see a balanced 
    distribution over phases, indicating that boundaries, and their 
    deep feature representations, are uniformly distributed in orientation.}
    \label{fig:layer1}
\vspace{-1em}
\end{figure}
The real parts of the filters, from the first layer of the 
boundary-detection-trained H-Net, are shown in Figure~\ref{fig:layer1}. 
They are aligned at zero phase ($\beta=0$) for ease of viewing. 
Since the network is trained on zero-mean, unit variance, normalized 
color images, the weights do not have the natural colors we would see 
in real-world images. Nonetheless, there is useful information we can 
glean from inspecting these. Most 1\textsuperscript{st} layer filters
detect color boundaries, there are no blank filters as one usually sees 
in CNNs, and there are few reoriented copies. We also see from the
phase histograms that all phases are utilized by filters throughout 
the network, indicating full use of the phase information. This is
interesting, because it means that the model's parameters are being
used fully, with low redundancy, which we surmise comes from easier
optimization on the equivariant manifold.
\begin{figure}[b]
\vspace{-1em}
	\includegraphics[width=\linewidth]{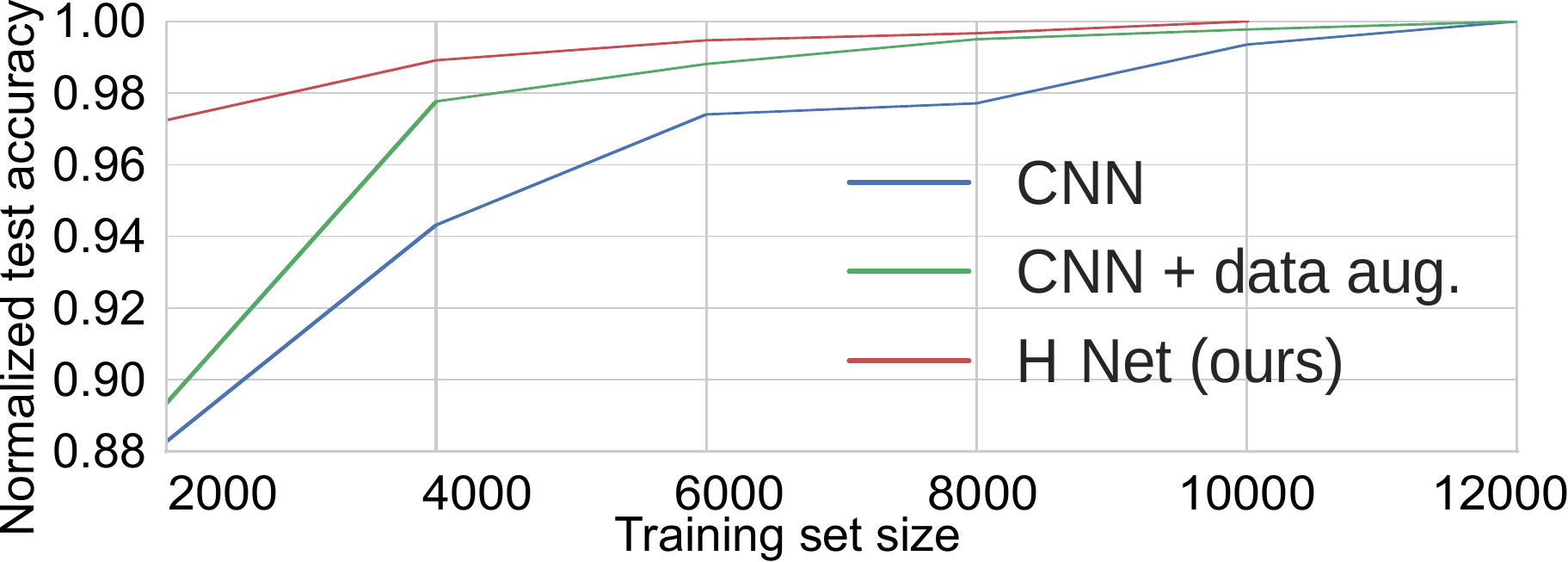}
    \caption{Data ablation study. On the rotated MNIST dataset, we experiment
    with test accuracy for varying sizes of the training set. We normalize the
    maximum test accuracy of each method to 1, for direct comparison of the falloff with
    training size. Clearly H-Nets are more data-efficient than regular CNNs,
    which need more data to discover rotational equivariance unaided.}
    \label{fig:DataAblation}
\end{figure}

\textbf{Data ablation}
Here we investigate H-nets data-efficiency. CNNs are massively data-hungry. Krizhevsky's landmark paper \cite{krizhevsky2012imagenet} used 60 million parameters, trained on 1.2 million $256\times256$ RGB images quantized to 256 bits and split between 1000 classes, for a total of 10 bits of information per weight. Even this vast amount of data was insufficient for training, and data augmentation was needed to improve results. We ran an experiment on the rotated MNIST dataset to show that with hard-baked rotation equivariance, we require less data than competing methods, which is indeed the case (see Figure~\ref{fig:DataAblation}). Interestingly, and predictably, regular CNNs trained with data augmentation still perform worse than H-Nets, because they only learn global invariance to rotation, rather than local equivariances at each layer.

\begin{figure}[b]
\vspace{-1em}
	\includegraphics[width=\linewidth]{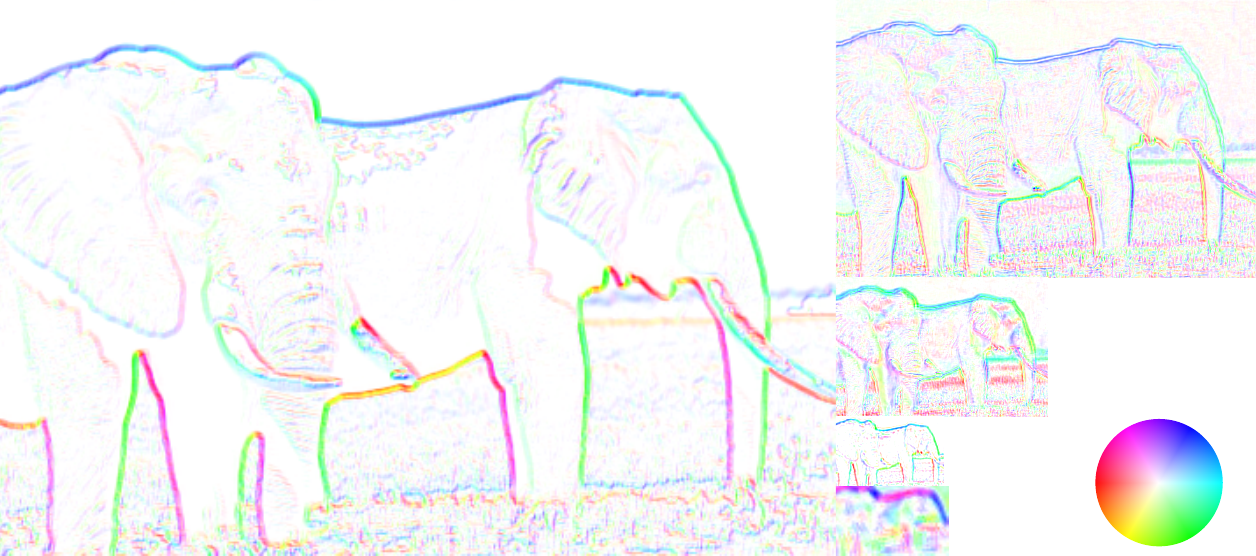}
    \caption{View best in color. 
    Orientated feature maps for the H-DSN. The color wheel shows
    orientation coding. Note that between layers boundary orientations
    are colored differently because each feature has a different $\beta$. 
    This visualization demonstrates the consistency
    of orientation within a feature map and across multiple layers.
    The images are taken from layers 2, 4, 6, 8, and 10 in a clockwise
    order from largest to smallest.}
    \label{fig:flow}
\end{figure}

\textbf{Feature maps}
\begin{figure}[t]
\begin{center}
	\includegraphics[width=0.8\linewidth]{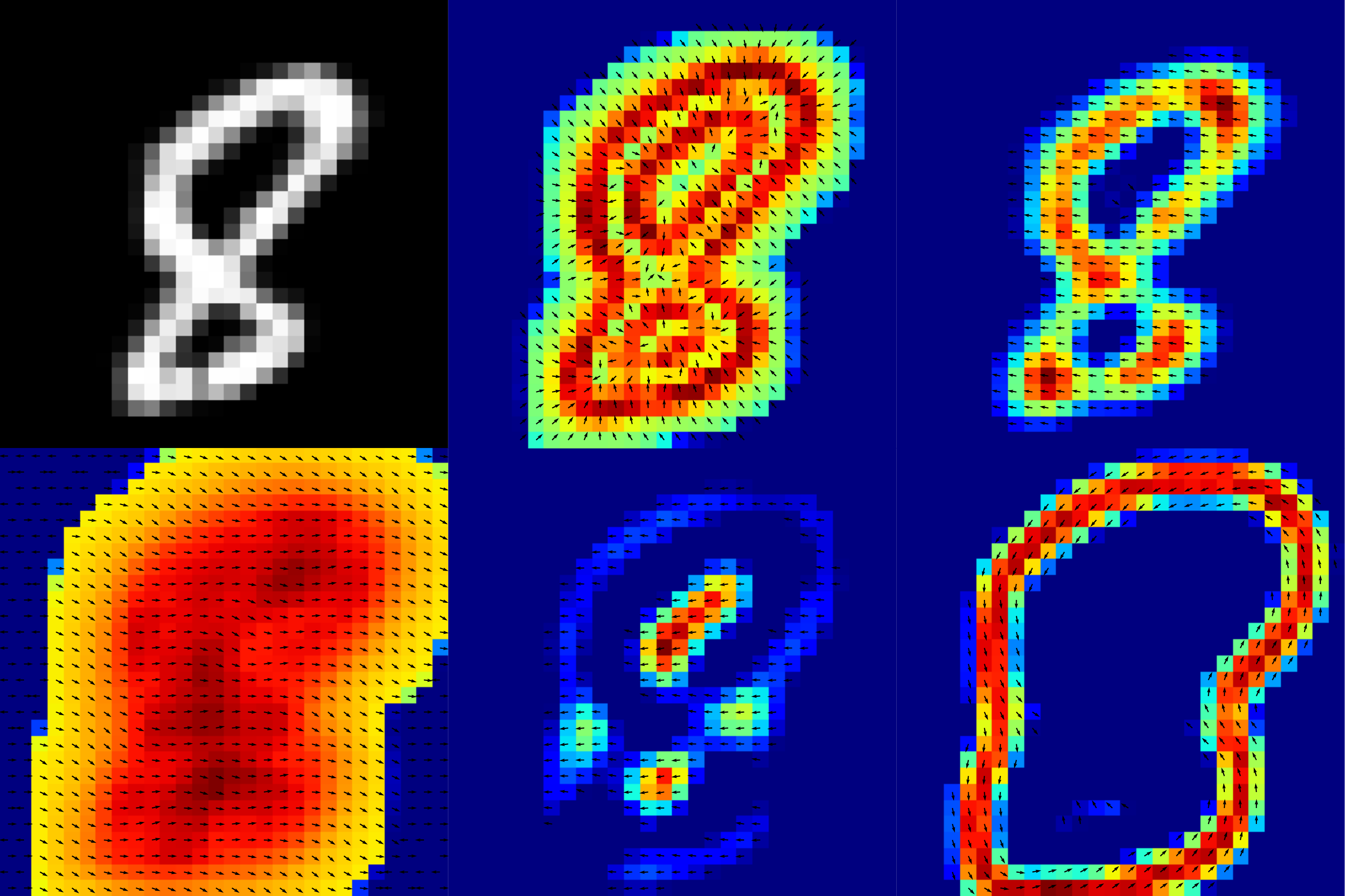}
\end{center}
\caption{Feature maps from the MNIST network. The arrows display phase, and the colors display magnitude information (jet color scheme). There are diverse features encoding edges, corners, whole objects, negative spaces, and outlines.}
\label{fig:feature_maps}
\vspace{-1em}
\end{figure}
We visualize feature maps in the lower layers of an MNIST trained 
H-Net (see Figure~\ref{fig:feature_maps}). 
For given input, we see the feature maps encode very 
complicated structures. Left to right, we see the H-Net 
learns to detect edges, corners, object presence, negative space, 
and outlines of objects. We perform this for the BSD500 trained
H-DSN (see Figure~\ref{fig:flow}). It shows equivariance
is preserved through to the deepest feature maps.
It also highlights the compact representation of feature presence and pose,
which regular CNNs cannot do.
\section{Conclusions}
We presented a convolutional neural network that is locally equivariant to patch-wise translation and, for the first time, to continuous $360^\circ$-rotation. We achieved this by restricting the filters to circular harmonics, essentially hard-baking rotation into the architecture. This can be implanted onto other architectures too. The use of circular harmonics pays dividends in that we receive full rotational equivariance using few parameters. This leads to good generalization, even with less (or less augmented) training data. The only disadvantage we've seen so far is the higher per-filter computational cost, but our guidance for network design balances that cost against the more expressive representation. The better interpretability of the feature maps is a bonus, because we know how they transform under input image rotations. We applied our network to the problem of classifying rotated-MNIST, setting a new state-of-the-art. We also applied our network to boundary detection, again achieving state-of-the-art results, for non-pretrained networks. We have shown that $360^\circ$-rotational equivariance is both possible and useful. Our TensorFlow\texttrademark   implementation is available on the project website. 

\textbf{Future work}
Extension of this work could involve hard-baking yet more transformations into the equivariance properties of the Harmonic Network, possibly extending to 3D. This will allow yet more expressibility in network representations, extending the benefits we have seen afforded by rotation equivariance to a larger class of models and applications.

\textbf{Acknowledgements}
Support is from Fight for Sight UK, a Microsoft Research PhD Scholarship, EPSRC Nature Smart Cities EP/K503745/1 and ENGAGE EP/K015664/1.

{\small
\bibliographystyle{ieee}
\bibliography{SteerBib}
}

\newpage
\onecolumn
\appendix
%%%%%%%%% TITLE
\begin{center}
\Large
\textbf{Supplementary Material}
\end{center}

%%%%%%%%% ABSTRACT
\begin{abstract}
	We include some proofs and derivations of the rotational equivariance
    properties of the circular harmonics, along with a demonstration of how
    we calculate the number of parameters for various network architectures.
\end{abstract}

%%%%%%%%% BODY TEXT

\section{Equivariance properties}
In Section 3.2 we mentioned that cross-correlation with the circular harmonics
is a $360^\circ$-rotation equivariant feature transform. Here we provide the proof,
and some of the properties mentioned in \bb{Arithmetic and Equivariance Condition}.

\subsection{Equivariance of the Circular Harmonics}
We are interested in proving that there exists a filter $\bb{W}_m$, such that cross-correlation of $\bb{F}$ with $\bb{W}_m$ yields a rotationally equivariant feature map. The proof requires us to introduce two different kinds of transformation: rotation $\mc{R}$ and translation $\mc{T}$. To simplify the math, we use vector notation, so the spatial domain of the filter/image is $\mbb{R}^2$. We write the filter as $\bb{W}_m(\bb{x})$ and image as $\bb{F}(\bb{x})$ for $\bb{x}\in\mbb{R}^2$. We define the transformation operators $\mc{R}_{\theta}$ and $\mc{T}_{\bb{t}}$, such that $\mc{R}_{\theta}\bb{F} = \bb{F}(\bb{R}_{-\theta}\bb{x})$ and $\mc{T}_{\bb{t}}\bb{F} = \bb{F}(\bb{x}-\bb{t})$, where $\bb{R}_{\theta}$ is a 2D rotation matrix for a $\theta$ counter-clockwise rotation. We introduce rotational cross-correlation $\star$. This is defined as
\begin{align}
	[\bb{W}_m \star \bb{F}] &= \int_{\Phi} \int_{R} \bb{W}_m(r \bb{R}_{\phi}\hat{\bb{x}}) \bb{F}(r\bb{R}_{\phi}\hat{\bb{x}}) \, \dd r \dd \phi,
\end{align}
where we have used the decomposition $\bb{x}=r\hat{\bb{x}}$, with $r=\|\bb{x}\|_2 \geq 0$ and $\hat{\bb{x}} = \bb{x}/r$. The rotational cross-correlation is performed about the origin of the image. If we rotate the image, then we have
\begin{align}
	[\bb{W}_m \star \mc{R}_{\theta}\bb{F}] &= \int_{\Phi} \int_{R} \bb{W}_m(r \bb{R}_{\phi}\hat{\bb{x}}) \bb{F}(r\bb{R}_{\phi}\bb{R}_{-\theta}\hat{\bb{x}}) \, \dd r \dd \phi \\
    &= \int_{\Phi} \int_{R} \bb{W}_m(r \bb{R}_{\phi}\hat{\bb{x}}) \bb{F}(r\bb{R}_{\phi-\theta}\hat{\bb{x}}) \, \dd r \dd \phi \\
    &= \int_{\Phi} \int_{R} \bb{W}_m(r \bb{R}_{\phi'+\theta}\hat{\bb{x}}) \bb{F}(r\bb{R}_{\phi'}\hat{\bb{x}}) \, \dd r \dd \phi'.
\end{align}
If we define $\bb{W}_m(\bb{x}) = \bb{W}_m(r\hat{\bb{x}}) = R(r)e^{i(m\phi + \beta)}$, where $\phi = \angle \hat{\bb{x}}$, then
\begin{align}
	[\bb{W}_m \star \mc{R}_{\theta}\bb{F}] &= \int_{\Phi} \int_{R} \bb{W}_m(r \bb{R}_{\phi'+\theta}\hat{\bb{x}}) \bb{F}(r\bb{R}_{\phi'}\hat{\bb{x}}) \, \dd r \dd \phi' \\
    &= \int_{\Phi} \int_{R} R(r)e^{i(m(\phi' + \theta) + \beta)} \bb{F}(r\bb{R}_{\phi'}\hat{\bb{x}}) \, \dd r \dd \phi' \\
    &= e^{im\theta}\int_{\Phi} \int_{R} R(r)e^{i(m\phi' + \beta)} \bb{F}(r\bb{R}_{\phi'}\hat{\bb{x}}) \, \dd r \dd \phi' \\
    & = e^{im\theta}[\bb{W}_m \star \bb{F}] \label{eq:equi}.
\end{align}
And so rotational cross-correlation is rotationally equivariant about the origin of rotation. In the next part, we build up to a result needed for proving the chained cross-correlation result.
\paragraph{Cross-correlation about $\bb{t}$}
To perform the rotational cross-correlation about another point $\bb{t}$, we first have to translate the image such that $\bb{t}$ is the new origin, so $F_{\bb{t}}(\bb{x}) = \bb{F}(\bb{x}-\bb{t})$, then perform the rotational cross-correlation, so
\begin{align}
	[\bb{W}_m \star \mc{T}_{\bb{t}}\bb{F}] &= [\bb{W}_m \star \bb{F}_{\bb{t}}] \\
    &=  \int_{\Phi} \int_{R} \bb{W}_m(r\bb{R}_{\phi}\hat{\bb{x}}) \bb{F}_{\bb{t}}(r\bb{R}_{\phi}\hat{\bb{x}}) \, \dd r_{x} \dd \phi \\
    &= \int_{\Phi} \int_{R} \bb{W}_m(r\bb{R}_{\phi}\hat{\bb{x}}) \bb{F}(r\bb{R}_{\phi}\hat{\bb{x}} - \bb{t}) \, \dd r_{x} \dd \phi.
\end{align}
\paragraph{Cross-correlation about $\bb{t}$ with rotated $\bb{F}$ about $\bb{t}$}
In general, for every $\bb{t}$ this expression returns a different value. The response of a $\theta$-rotated image about $\bb{t}$ is then
\begin{align}
	[\bb{W}_m \star \mc{R}_{\theta}\mc{T}_{\bb{t}}\bb{F}] &= [\bb{W}_m \star \mc{R}_{\theta}\bb{F}_{\bb{t}}] \\
    &= \int_{\Phi} \int_{R} \bb{W}_m(r\bb{R}_{\phi}\hat{\bb{x}}) \bb{F}_{\bb{t}}(r\bb{R}_{-\theta}\bb{R}_{\phi}\hat{\bb{x}}) \, \dd r_{x} \dd \phi \\
    &= \int_{\Phi} \int_{R} \bb{W}_m(r\bb{R}_{\phi}\hat{\bb{x}}) \bb{F}_{\bb{t}}(r\bb{R}_{\phi-\theta}\hat{\bb{x}}) \, \dd r_{x} \dd \phi \\
    &= \int_{\Phi} \int_{R} \bb{W}_m(r\bb{R}_{\phi'+\theta}\hat{\bb{x}}) \bb{F}_{\bb{t}}(r\bb{R}_{\phi'}\hat{\bb{x}}) \, \dd r_{x} \dd \phi' \\
    &= e^{im\theta} \int_{\Phi} \int_{R_z} R(r_z)e^{i(m\phi' + \beta)} \bb{F}_{\bb{t}}(r\bb{R}_{\phi'}\hat{\bb{x}}) ) \, \dd r \dd \phi' \\
    &= e^{im\theta}[\bb{W}_m \star \mc{T}_{\bb{t}}\bb{F}].
\end{align}
\paragraph{Cross-correlation about $\bb{t}$ with rotated $\bb{F}$ about origin}
Say we wish to perform the rotational cross-correlation about a point $\bb{t}$, when the image has been rotated about the origin. Denoting $\bb{F}^{\theta} = \mc{R}_{\theta}\bb{F}$, then the response is
\begin{align}
	[\bb{W}_m \star \mc{T}_{\bb{t}}\mc{R}_{\theta}\bb{F}] &= [\bb{W}_m \star \mc{T}_{\bb{t}}\bb{F}^{\theta}] \\
    &= \int_{\Phi} \int_{R} \bb{W}_m(r\bb{R}_{\phi}\hat{\bb{x}}) \bb{F}^{\theta}(r_x\bb{R}_{\phi}\hat{\bb{x}} - \bb{t}) \, \dd r_x \dd \phi \\
    &= \int_{\Phi} \int_{R} \bb{W}_m(r\bb{R}_{\phi}\hat{\bb{x}}) \bb{F}(r_x\bb{R}_{-\theta}\bb{R}_{\phi}\hat{\bb{x}} - \bb{R}_{-\theta}\bb{t}) \, \dd r_x \dd \phi \\
    &= \int_{\Phi} \int_{R} \bb{W}_m(r\bb{R}_{\phi}\hat{\bb{x}}) \bb{F}(r_x\bb{R}_{\phi-\theta}\hat{\bb{x}} - \bb{R}_{-\theta}\bb{t}) \, \dd r_x \dd \phi \\
    &= \int_{\Phi} \int_{R} \bb{W}_m(r\bb{R}_{\phi'+\theta}\hat{\bb{x}}) \bb{F}(r_x\bb{R}_{\phi'}\hat{\bb{x}} - \bb{R}_{-\theta}\bb{t}) \, \dd r_x \dd \phi' \\
    &= e^{im\theta} \int_{\Phi} \int_{R} \bb{W}_m(r\bb{R}_{\phi'}\hat{\bb{x}}) \bb{F}(r_x\bb{R}_{\phi'}\hat{\bb{x}} - \bb{R}_{-\theta}\bb{t}) \, \dd r_x \dd \phi' \\
    &= e^{im\theta}[\bb{W}_m \star \mc{T}_{\bb{R}_{-\theta}\bb{t}}\bb{F}].
\end{align}

Thus we see that cross-correlation of the rotated signal $\bb{F}^{\theta}$ with the circular harmonic filter $\bb{W}_m = R(r)e^{i(m\phi+\beta)}$ is equal to the response at zero rotation $[\bb{W} \star \bb{F}]$, multiplied by a complex phase shift $e^{im\theta}$. In the notation of the paper, we denote this multiplication by $e^{im\theta}$ as $\psi_m^{\theta} [\bullet] = e^{im\theta} \cdot \bullet$. Thus cross-correlation with $\bb{W}_m$ yields a rotationally equivariant feature mapping.

\subsection{Properties}
\subsubsection{Chained cross-correlation}
We claimed in \bb{Arithmetic and Equivariance Condition}, that the rotation order of a feature map resulting from chained cross-correlations is equal to the sum of the the rotation orders of the filters in the chain. We prove this for a chain of two filters, and the rest follows by induction. Consider taking a $\theta$-rotated image $\bb{F}$ about the origin, then cross-correlating it with a filter $\bb{W}_m$ as every point in the image plane $\bb{t}\in\mbb{R}^2$, followed by cross-correlation with $\bb{W}_n$ as a point $\bb{s}\in\mbb{R}^2$. We already know that the response to the rotation is $[\bb{W}_m \star \mc{T}_{\bb{t}}\mc{R}_{\theta}\bb{F}] = e^{im\theta}[\bb{W}_m \star \mc{T}_{\bb{R}_{-\theta}\bb{t}}\bb{F}]$, for all rotations $\theta$ of the input and all points $\bb{t}$ in the response plane, so we can write the chained convolution as 
\begin{align}
	[\bb{W}_n \star \mc{T}_{\bb{s}}[\bb{W}_m \star \mc{T}_{\bb{t}}\mc{R}_{\theta}\bb{F}]] &= [\bb{W}_n \star \mc{T}_{\bb{s}}e^{im\theta}[\bb{W}_m \star \mc{T}_{\bb{R}_{-\theta}\bb{t}}\bb{F}]] \\
    &= e^{im\theta} \left [\bb{W}_n \star \mc{T}_{\bb{s}}[\bb{W}_m \star \mc{T}_{\bb{R}_{-\theta}\bb{t}}\bb{F}]] \right]
\end{align}
We have used the property that the cross-correlation is linear and that we may pull the scalar factor $e^{im\theta}$ outside. If we write $\bb{G}(\bb{t}) = [\bb{W}_m \star \mc{T}_{\bb{t}}\bb{F}]$ then $[\bb{W}_m \star \mc{T}_{\bb{R}_{-\theta}\bb{t}}\bb{F}] = \bb{G}(\bb{R}_{-\theta}\bb{t}) = [\mc{R}_{\theta}\bb{G}](\bb{t})$, so
\begin{align}
	[\bb{W}_n \star \mc{T}_{\bb{s}}[\bb{W}_m \star \mc{T}_{\bb{t}}\mc{R}_{\theta}\bb{F}]] &= e^{im\theta} \left [\bb{W}_n \star \mc{T}_{\bb{s}}[\bb{W}_m \star \mc{T}_{\bb{R}_{-\theta}\bb{t}}\bb{F}]] \right] \\
    &= e^{im\theta}[\bb{W}_n \star \mc{T}_{\bb{s}}\mc{R}_{\theta}\bb{G}] \\
    &= e^{im\theta}  e^{in\theta}\left [\bb{W}_n \star \mc{T}_{\bb{R}_{-\theta}\bb{s}}\bb{G} \right ].
\end{align}
Thus we see that the chained cross-correlation results in a summation of the rotation orders of the individual filters $\bb{W}_m$ and $\bb{W}_n$. Setting $\bb{s}=\bb{0}$, such that we evaluate the cross-correlation at the center of rotation, we regain an equation similar to \ref{eq:equi}.

\subsubsection{Magnitude nonlinearities}
Point-wise nonlinearities acting on the magnitude of a feature map maintain rotational 
equivariance. Consider that we have a point on a feature map of rotation order $m$, which 
we can write as $Fe^{im\theta}$, where $F \geq 0$ is the magnitude of the feature map and 
$e^{im\theta}$ is the phase component. The output of the nonlinearity $g:\mbb{R}_+\to\mbb{R}$ 
is
\begin{align}
	g(Fe^{im\theta}) = g(F)e^{im\theta},
\end{align}
since $g$ only acts on magnitudes. Since for fixed $F$ the output is a function of $m$
and $\theta$ only, the point-wise magnitude-acting nonlinearity preserves rotational
equivariance.

\subsubsection{Summation of feature maps}
The summation of feature maps of the same rotation order is a new feature map of the
same rotation order. Consider two feature maps $\bb{F}_1$ and $\bb{F}_2$ of rotation
order $m$. Summation is a pointwise operation, so we only consider two corresponding 
points in the feature maps, which we denote $F_1e^{i(m\theta + \beta_1)}$ and 
$F_2e^{i(m\theta + \beta_2)}$, where $\beta_1$ and $\beta_2$ are phase offsets. The
sum is
\begin{align}
	F_1e^{i(m\theta + \beta_1)} + F_2e^{i(m\theta + \beta_2)} = e^{im\theta}\left ( F_1e^{i\beta_1} + F_2e^{i\beta_2} \right ),
\end{align}
which for fixed $F_1, F_2, \beta_1, \beta_2$ is a function of $m$ and $\theta$ only 
and so also rotationally equivariant with order $m$.

\newpage
\section{Number of parameters}
Here we list a break down of how we computed the number of parameters for the 
various network architectures in the experiments section. The networks architectures
used are in Figure \ref{fig:networks}. Red boxes are cross-correlations, blue
boxes are pooling (average for H-Nets, max for regular CNNs), green boxes are 
$1\times 1$-cross-correlations.
\begin{figure}[h!]
	\centering
	\includegraphics[width=0.5\linewidth]{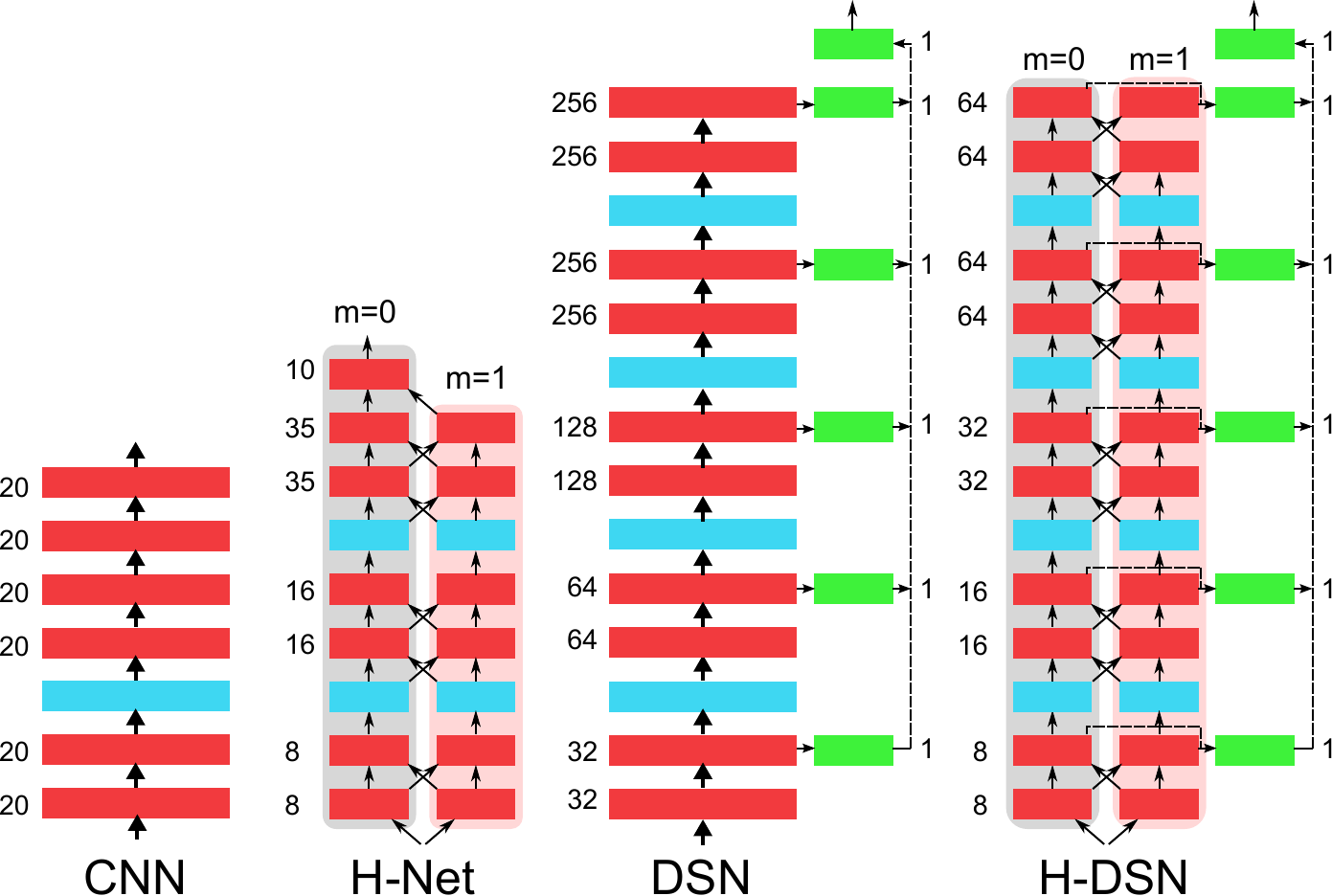}
    \caption{Networks used}
    \label{fig:networks}
\end{figure}

\subsection{Standard CNN}
For a standard CNN layer with $i$ input channels and $o$ output channels, and
$k\times k$ sized weights, the number of learnable parameters is $iok^2$.
Since there is one bias per output layer, this increases to $iok^2 + o$. If using
batch normalization, then there is an extra per-channel scaling factor, which 
increases the number of learnable parameters to $iok^2 + 2o$. The standard CNN
for the rotated MNIST experiments has 6 layers of $3\times 3$ cross-correlations, and 
1 layer of $4\times4$-cross-correlations, with 20 feature maps per layer and 3 batch 
normalization layers so the number of learnable parameters is 21570. The calculations
are shown in Table \ref{tab:cnn}.
\begin{table}[h]
  \begin{center}
    \begin{tabular}{|l|c|c|c|c|}
    \hline
    Layer	& Weights		& Batch Norm/Bias	& \#Params\\
    \hline\hline
    1 	& $3\cdot3\cdot1\cdot20$	& $20$			&	200\\
    2 	& $3\cdot3\cdot20\cdot20$	& $2\cdot 20$	& 	3640\\
    3 	& $3\cdot3\cdot20\cdot20$	& $20$			&	3620\\
    4 	& $3\cdot3\cdot20\cdot20$	& $2\cdot 20$	&	3640\\
    5 	& $3\cdot3\cdot20\cdot20$	& $20$			&	3620\\
    6 	& $3\cdot3\cdot20\cdot20$	& $2\cdot 20$	&	3640\\
    7 	& $4\cdot4\cdot20\cdot10$	& $10$			&	3210\\
    \hline
    \text{Total}	&				&				&	21570\\
    \hline
    \end{tabular}
  \end{center}
  \caption{Number of parameters for a regular CNN.}
  \label{tab:cnn}
\end{table}

\subsection{Harmonic networks}
\begin{figure}[t]
\centering
	\includegraphics[width=0.7\linewidth]{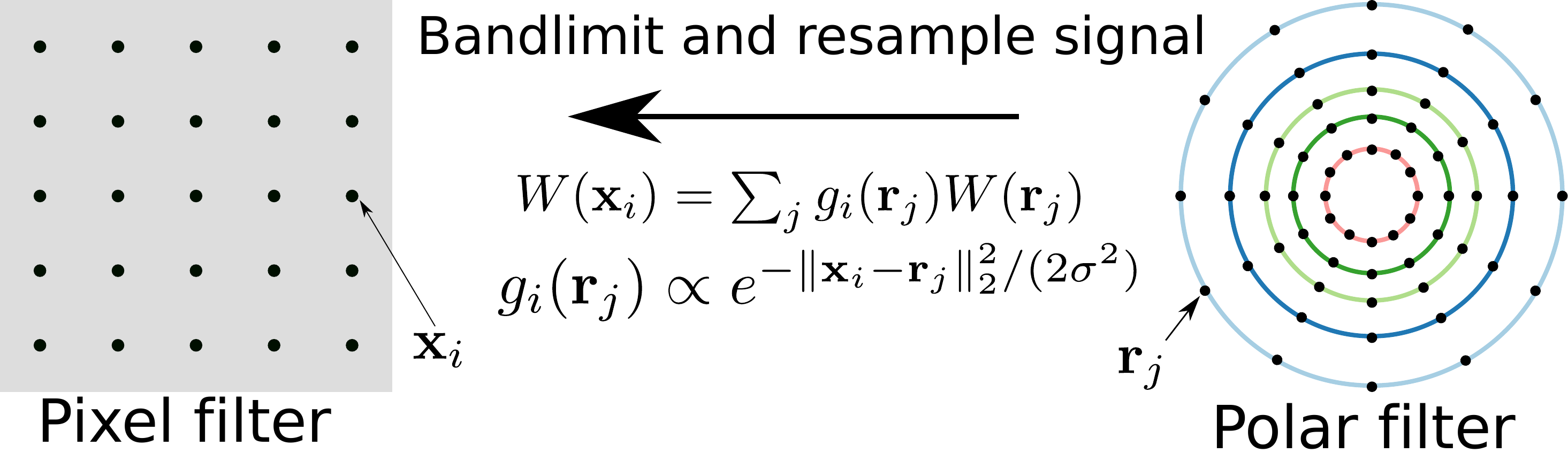}
    \caption{Each radius has a single learnable weight. Then there is a bias for the whole filter.}
    \label{fig:radial_profile}
\end{figure}

The learnable parameters of a Harmonic Network are the radial profile and the
the per-filter phase offset. For a $k\times k$ filter, the number of radial
profile elements is equal to the number of rings of equal distance from the
center of the filter. For example, consider the Figure \ref{fig:radial_profile},
which is an excerpt from the main paper. This is a $5\times 5$ filter, with
6 rings of equal distance from the center of the filter (the smallest ring 
is just a single point). So this filter has 6 radial profile terms and 1 
phase offset to learn. This contrasts with a regular filter, which would have
25 learnable parameters. Note, that for filters with rotation order $m\neq 0$,
the center pixel of the filter is in fact always zero, and so for a $5\times 5$
rotation order $m\neq 0$ filter, the number of radial profile terms is $6-1=5$.
So for the H-Net in the main paper with $5\times 5$ filters and batch normalization
in layers 2, 4, \& 6, the number of learnable parameters is 33347. The calculations
are in Table \ref{tab:h_net}. Note that the final layer contains just one set 
of biases and no phase offsets. A similar set of calculations can be performed 
for the deeply supervised networks.

\begin{table}[h]
  \begin{center}
    \begin{tabular}{|l|c|c|c|c|}
    \hline
    Layer	& $m=0$		& $m=1$		& Batch Norm/Bias	& \#Params\\
    \hline\hline
    1 	& $6\cdot1\cdot8 + 1\cdot8$		& $5\cdot1\cdot8 + 1\cdot8$		& $2\cdot 8$	&	120\\
    2 	& $6\cdot8\cdot8 + 8\cdot8$		& $5\cdot8\cdot8 + 8\cdot8$		& $2\cdot 16$	& 	864\\
    3 	& $6\cdot8\cdot16 + 8\cdot16$	& $5\cdot8\cdot16 + 8\cdot16$	& $2\cdot 16$	&	1696\\
    4 	& $6\cdot16\cdot16 + 16\cdot16$	& $5\cdot16\cdot16 + 16\cdot16$	& $2\cdot 32$	&	3392\\
    5 	& $6\cdot16\cdot35 + 16\cdot35$	& $5\cdot16\cdot35 + 16\cdot35$	& $2\cdot 35$	&	7350\\
    6 	& $6\cdot35\cdot35 + 35\cdot35$	& $5\cdot35\cdot35 + 35\cdot35$	& $2\cdot 70$	&	16065\\
    7 	& $6\cdot35\cdot10$				& $5\cdot35\cdot10$				& $10$			&	3860\\
    \hline
    \text{Total}	&					&								&				&	33347\\
    \hline
    \end{tabular}
  \end{center}
  \caption{Number of parameters for H-Net.}
  \label{tab:h_net}
\end{table}

\end{document}